\documentclass[conference]{IEEEtran}
\IEEEoverridecommandlockouts
\usepackage{cite}
\usepackage{amsmath,amssymb,amsfonts}
\usepackage{graphicx, enumitem}
\usepackage{textcomp}
\usepackage[ruled,vlined,linesnumbered]{algorithm2e}
\usepackage[table]{xcolor}
\usepackage{tikz}
\usepackage{bm}
\usepackage{booktabs}
\usepackage{multirow}
\usepackage{url}

\newcommand{\tikzcircle}[2][red,fill=red]{\tikz[baseline=-0.5ex]\draw[#1,radius=#2] (0,0) circle ;}

\newcolumntype{L}[1]{>{\raggedright\let\newline\\\arraybackslash\hspace{0pt}}m{#1}}
\newcolumntype{C}[1]{>{\centering\let\newline\\\arraybackslash\hspace{0pt}}m{#1}}
\newcolumntype{R}[1]{>{\raggedleft\let\newline\\\arraybackslash\hspace{0pt}}m{#1}} 

\def\BibTeX{{\rm B\kern-.05em{\sc i\kern-.025em b}\kern-.08em
    T\kern-.1667em\lower.7ex\hbox{E}\kern-.125emX}}

\begin{document}

\title{Multifactorial Cellular Genetic Algorithm (MFCGA): Algorithmic Design, Performance Comparison and Genetic Transferability Analysis}

\author{
\IEEEauthorblockN{Eneko Osaba\IEEEauthorrefmark{2}\IEEEauthorrefmark{1},
Aritz D. Martinez\IEEEauthorrefmark{2}\IEEEauthorrefmark{1},
Jesus L. Lobo\IEEEauthorrefmark{2},
Javier Del Ser\IEEEauthorrefmark{2}\IEEEauthorrefmark{3} and
Francisco Herrera\IEEEauthorrefmark{4}}
\IEEEauthorblockA{\IEEEauthorrefmark{2}TECNALIA, Basque Research and Technology Alliance (BRTA), 48160 Derio, Bizkaia, Spain\\
Email: [eneko.osaba, aritz.martinez, jesus.lopez, javier.delser]@tecnalia.com}
\IEEEauthorblockA{\IEEEauthorrefmark{3}University of the Basque Country (UPV/EHU), 48013 Bilbao, Bizkaia, Spain\\}
\IEEEauthorblockA{\IEEEauthorrefmark{4}DaSCI Andalusian Institute of Data Science and Computational Intelligence. University of Granada. 18071 Granada, Spain\\}
\IEEEauthorblockA{\IEEEauthorrefmark{1} Corresponding authors. These authors have equally contributed to the work presented in this paper.}}
\maketitle

\begin{abstract}
Multitasking optimization is an incipient research area which is lately gaining a notable research momentum. Unlike traditional optimization paradigm that focuses on solving a single task at a time, multitasking addresses how multiple optimization problems can be tackled simultaneously by performing a single search process. The main objective to achieve this goal efficiently is to exploit synergies between the problems (tasks) to be optimized, helping each other via knowledge transfer (thereby being referred to as Transfer Optimization). Furthermore, the equally recent concept of Evolutionary Multitasking (EM) refers to multitasking environments adopting concepts from Evolutionary Computation as their inspiration for the simultaneous solving of the problems under consideration. As such, EM approaches such as the Multifactorial Evolutionary Algorithm (MFEA) has shown a remarkable success when dealing with multiple discrete, continuous, single-, and/or multi-objective optimization problems. In this work we propose a novel algorithmic scheme for Multifactorial Optimization scenarios -- the Multifactorial Cellular Genetic Algorithm (MFCGA) -- that hinges on concepts from Cellular Automata to implement mechanisms for exchanging knowledge among problems. We conduct an extensive performance analysis of the proposed MFCGA and compare it to the canonical MFEA under the same algorithmic conditions and over 15 different multitasking setups (encompassing different reference instances of the discrete Traveling Salesman Problem). A further contribution of this analysis beyond performance benchmarking is a quantitative examination of the genetic transferability among the problem instances, eliciting an empirical demonstration of the synergies emerged between the different optimization tasks along the MFCGA search process.
\end{abstract}


\section{Introduction} \label{sec:intro}

Inspired by the roots of Transfer Learning \cite{pan2009survey} and Multitask Learning \cite{caruana1997multitask}, Transfer Optimization is a relatively new knowledge field within the wider area of optimization, which is attracting great attention within the community in recent years \cite{ong2016evolutionary}. The main idea is to exploit what has been learned for optimizing one given optimization problem, toward tackling another related or unrelated problem. Due to its relative youth, efforts devoted by the scientific community for advancing over this emerging research area are considerably fewer than those dedicated to Transfer and Multitask Learning, which address a similar problem for Machine Learning tasks. It has not been until recently when the transferability of knowledge among optimization problems have become a research priority, mainly due to the increasing complexity and scales of optimization problems, and the subsequent need for harnessing knowledge acquired beforehand.  

Three different algorithmic categories can be identified in the Transfer Optimization panorama \cite{gupta2017insights}. The first one is \textit{sequential transfer}, in which optimization problems (tasks) are solved in a sequential fashion under the assumption that for optimizing a new problem/instance, the knowledge acquired when solving previous tasks is used as external information \cite{feng2015memes}. The second class is the so-called \textit{multitasking}, which tackles different tasks of equal priority in a simultaneous way by dynamically exploiting existing synergies and complementarities among problems \cite{gupta2016genetic,wen2017parting}. The last category is referred to as \textit{multiform optimization}, which aims at the resolution of a single task through the use of alternative problem formulations, which are solved in a simultaneous way. As can be observed in the literature background, in all these three categories the correlation among problem instances or tasks is crucial for positively capitalizing on the transference of knowledge over the search \cite{gupta2015multifactorial}.

Among the above three categories, the most prolific one in the current literature is \textit{multitasking}. The research presented in this study is centered on this category. Being more specific, we focus on multitasking optimization through the perspective of Evolutionary Multitasking (EM, \cite{ong2016towards}). In short, EM embraces the concepts, operators and search strategies conceived within Evolutionary Computation for simultaneously solving several problems at a time \cite{back1997handbook,del2019bio}. As such, EM underlies Multifactorial Optimization (MFO, \cite{gupta2015multifactorial}), a particular realization of this paradigm that has demonstrated its potential in different environments encompassing continuous, discrete and multi-objective optimization problems \cite{wang2019evolutionary,gong2019evolutionary,yu2019multifactorial,gupta2016multiobjective}. In the current community, the majority of contributions around MFO gravitate on the Multifactorial Evolutionary Algorithm (MFEA, \cite{gupta2015multifactorial}), or variants of this algorithm.

Bearing this background in mind, this work presents a new MFO approach coined as Multifactorial Cellular Genetic Algorithm (MFCGA). We take a step further over the state of the art by elaborating on several research directions:
\begin{itemize}[leftmargin=*]
\item We introduce a new efficient meta-heuristic scheme for MFO that relies on the foundations of Cellular Automata and Celular Genetic Algorithms (cGAs, \cite{manderick1989fine}) for controlling the mating process among different species (problems). Moreover, the search strategy of the proposed MFCGA solver favors the exploration and quantitative examination of synergies among the problems being solved, providing a sort of explainability interface for understanding the interactions between problems. To the best of our knowledge, MFCGA is the first cGA for being applied to EM, and to the wider Transfer Optimization domain.

\item We conduct an extensive experimentation focused on the well-known Traveling Salesman Problem (TSP, \cite{lawler1985traveling}). Specifically, we compare the performance of our proposed solver with those obtained by the canonical MFEA, with the firm intention of demonstrating that our solver is a promising alternative to face MFO scenarios. To do that, we employ 8 different TSP instances, which have been used to generate 15 different scenarios. We also examine the genetic transferability among the used TSP instances, which poses a valuable addition to the state of the art \cite{zhou2018study,gupta2016landscape}, and provides useful insights for further research.
\end{itemize}

The rest of the paper is organized as follows: Section \ref{sec:back} presents a brief overview of the background related to Evolutionary Multitasking, MFEA and cGAs. Next, in Section \ref{sec:MFCGA}, we describe the main characteristics of our proposed MFCGA in detail. Experimental results obtained with the developed method are discussed in Section \ref{sec:exp}, along with a description of the benchmark and the experimental setup under consideration. Finally, Section \ref{sec:conc} concludes the paper by drawing conclusions and outlining future research lines.

\section{Background} \label{sec:back}

As stated above, this section is devoted to providing a brief background about the three main concepts addressed in this paper: EM and MFO (Section \ref{sec:back_EM}), MFEA (Section \ref{sec:MFEA}), and the area of cGAs (Section \ref{sec:back_CGA}).

\subsection{Evolutionary Multitasking and Multifactorial Optimization} \label{sec:back_EM}

In contrast to sequential transfer, in which a single optimization task is addressed at a time, multitasking focuses on simultaneously addressing several tasks. While sequential transfer optimization seeks an unidirectional transfer of knowledge from previously completed tasks to new ones, multitasking is characterized by omnidirectional knowledge transfer among tasks, pursuing a more synergistic completion of the tasks under consideration \cite{gupta2017insights}.

Within this landscape, EM has emerged as a promising paradigm for dealing with simultaneous transfer optimization scenarios. Two main characteristics motivated the first formulation of the EM paradigm. The first feature is the inherent parallelism offered by a population of individuals, which allows for efficient computational means to deal with concurrent optimization tasks faced simultaneously. It is precisely this simultaneous treatment what permits latent relationships between problems to be automatically harnessed during the search process \cite{ong2016evolutionary}. The second interesting characteristic is that the constant transfer of genetic material along the evolutionary search allows all tasks to benefit each other, even for tasks that are not strongly correlated \cite{louis2004learning,gupta2015multifactorial}.

It was not until late 2017 when the concept of EM was only formalized through the perspective of the MFO paradigm \cite{da2017evolutionary}. Firstly introduced in \cite{gupta2015multifactorial}, this incipient branch of the evolutionary computation field is grasping notable interest in terms of new algorithmic schemes, such as hybrid solvers \cite{xiao2019multifactorial}, multifactorial heuristic engines encompassing modern metaphors \cite{zheng2016multifactorial} or multi-population methods \cite{song2019multitasking} under the development of a novel multitasking multi-swarm optimization. Despite this recent upsurge of new MFO schemes, MFEA has dominated the knowledge stream of MFO since its conception.

Mathematically speaking, MFO can be formally described as an EM environment in which $K$ optimization tasks are simultaneously optimized. This environment is characterized in this way by the existence of multiple search spaces, each related to a single task. Assuming that all tasks are minimization problems, for the $k$-th task $T_k$ its objective function is characterized as $f_k : \Omega_k \rightarrow \mathbb{R}$, where $\Omega_k$ denotes the solution space of $T_k$. This being said, the main goal of MFO is to find a set of solutions $\{\mathbf{x}_1, \mathbf{x}_2,\dots,\mathbf{x}_K\}$ such that $\mathbf{x}_k = \arg \min_{\mathbf{x}\in\Omega_k} f_k(\mathbf{x})$. However, instead of tackling $K$ independent search processes in isolation, MFO pursues to find $\{\mathbf{x}_k\}_{k=1}^K$ by exploring a single, unified search space $\Omega^\prime$. Therefore, solutions $\mathbf{x}^\prime\in\Omega^\prime$ can be encoded and decoded to represent a task-specific solution $\mathbf{x}_k$ for any of the $K$ optimization tasks under consideration. 

Moreover, MFO is based on four different definitions, associated to each individual $\mathbf{x}_p^\prime\in\Omega^\prime$ within a $P$-sized population:

\textit{Definition 1 (Factorial Cost)}: the factorial cost $\Psi_k^p$ of a population member $\mathbf{x}_p^\prime$ is equal to the value of the fitness function for a given task $T_k$. Each individual counts with a list $\{\Psi_1^p,\Psi_2^p,\dots,\Psi_K^p\}$ of factorial costs, each related to an optimization task.

\textit{Definition 2 (Factorial Rank)}: the factorial rank $r_k^p$ of an individual $\mathbf{x}_p$ in a given task $T_k$ is the index of this individual within the population sorted in ascending order of $\Psi_k^p$. Each population member has a factorial rank list $\{r_1^p,r_2^p,\dots,r_K^p\}$.

\textit{Definition 3 (Scalar Fitness)}: the scalar fitness $\varphi^p$ of $\mathbf{x}_p^\prime$ is calculated by using the best factorial ranks over all the tasks, i.e., $\varphi^p = 1/ \left(\min_{k \in \{1...K\}}r_k^p\right)$. As will be exposed in Section \ref{sec:MFEA}, the scalar fitness is used for comparing individuals in MFEA.

\textit{Definition 4 (Skill Factor)}: The skill factor $\tau^p$ is the task in which $\mathbf{x}_p^\prime$ performs best, namely, $\tau^p = \arg \min_{k\in\{1,\ldots,K\}} r_k^p$. The skill factor plays a crucial role in MFEA by establishing which population members are selected for crossover.

The research activity around MFO and MFEA has been vibrant in the last few years. In \cite{yuan2016evolutionary} MFEA is applied to different discrete problems, such as the job shop scheduling problem and the TSP. This paper also introduces the unified discrete encoding strategy, which is embraced in this work. A similar study is proposed in \cite{zhou2016evolutionary}, where MFEA is put to practice to deal with vehicle routing problems. In \cite{gupta2016multiobjective}, a multiobjective variant of MFEA is proposed, proving its efficiency over continuous benchmark functions, as well as a real-world manufacturing process design problem. An interesting discrete MFEA is also developed in \cite{wang2019evolutionary} for Semantic Web Service Composition. An improved version of MFEA was proposed in \cite{gong2019evolutionary}, which endowed the algorithm with a dynamic resource allocating strategy. Likwewise, the enhanced MFEA presented in \cite{yu2019multifactorial} follows a similar philosophy by incorporating opposition-based learning. Further works around MFO and MFEA can be found in \cite{li2018multipopulation,zhou2019towards} and \cite{yu2019multifactorial}.

\subsection{Multifactorial Evolutionary Algorithm}\label{sec:MFEA}

MFEA is a recently proposed MFO method for solving EM problems using bio-cultural schemes of multifactorial inheritance \cite{gupta2015multifactorial}. The basic workflow of MFEA is depicted in Algorithm \ref{alg:classicMFEA}. For deeper details on the algorithmic operators, we refer reader to \cite{gupta2015multifactorial}. For simultaneously dealing with all optimizing tasks, MFEA has four cornerstone characteristics: unified solution representation, assortative mating, selective evaluation, and scalar fitness based selection:
\begin{itemize}[leftmargin=*]
\item The design of the representation strategy for $\mathbf{x}_p^\prime$ that yields the unified search space $\Omega^\prime$ is subject to the characteristics of the $K$ problems under consideration. Specifically for this work, the TSP is used as the family of benchmark problems for assessing the performance of both MFEA and the proposed MFCGA. For this reason, the well-known permutation encoding is used as the unified representation for $\mathbf{x}_p^\prime$ \cite{bierwirth1996permutation}. Following \cite{yuan2016evolutionary}, if $K$ TSP problems are to be simultaneously solved, and by denoting the size of each TSP problem $T_k$ (i.e. the number of \emph{cities}) as $D_k$, an individual $\mathbf{x}_p^\prime$ is encoded as a permutation of the integer set $\{1,2,\ldots, D_{max}\}$, where $D_{max}=\max_{k\in\{1,\ldots,K\}} D_k$, namely, the maximum problem size among the $K$ tasks. In this way, when $\mathbf{x}_p^\prime$ is to be evaluated for a task $T_k$ whose $D_k<D_{max}$, only integers lower than $D_k$ are considered for producing the argument solution $\mathbf{x}_k$ of $f_k(\cdot)$. These integers maintain the same order as in $\mathbf{x}_p^\prime$.

\item Assortative mating establishes that individuals prefer to interact with other mates belonging to similar cultural background \cite{gupta2015multifactorial}. Thus, as described in \cite{li2018multipopulation,gupta2016multiobjective,binh2018effective}, genetic operators used in the MFEA promote mating among individuals with the same skill factor $\tau^p$. We recommend consulting these papers for deeper details on how this breeding mechanism is implemented in MFEA.

\item Selective evaluation implies that each generated individual is measured only for one task, instead of evaluating it for every task. Specifically, the produced offspring is evaluated in task $T_{\tau_\ast^p}$, where $\tau_\ast^p$ is the skill factor of its parent (or the skill factor selected at random among the two parents of the offspring). This means that the factorial cost $\Psi_k^p$ is set to $\infty$ $\forall k\in\{1,\ldots,\tau_\ast^p-1,\tau_\ast^p+1,\ldots,K\}$.

\item Finally, the scalar fitness based selection is a survivor function similar to those used in basic Genetic Algorithms. In this case, MFEA is based on an elitist strategy, i.e. the best $P$ individuals in terms of scalar fitness $\sigma^p$ among those in the current population and the newly produced offspring survive for the next generation.
\end{itemize}
\begin{algorithm}[h!]
	\SetAlgoLined
	\DontPrintSemicolon
	Randomly generate a population of $P$ individuals\;
	Evaluate each generated individual for the $K$ problems\;
	Calculate the skill factor ($\tau^p$) of each individual $\mathbf{x}_p^\prime$\;
	\Repeat{termination criterion not reached}{
		Apply genetic operators on $P$ to get the offspring subpopulation $P_\ast$\;
		Evaluate the generated offspring for the best task $\tau_\ast^p$ of their parent(s)\;
		Combine $P$ and $P_\ast$ in intermediate population $Q$\;
		Update the scalar fitness $\varphi_k^p$ and skill factor $\tau^p$ for each individual in $Q$\;
		Build the next population by selecting the best $P$ individuals in $Q$ in terms of scalar fitness\;
	}
	Return the best individual for each task $T_k$\;
	\caption{Pseudocode of the canonical MFEA}
	\label{alg:classicMFEA}
\end{algorithm}

\subsection{Cellular Genetic Algorithm} \label{sec:back_CGA}

Briefly explained, cGAs are a sub-type of the canonical GAs in which the population is structured in a specific topology based on small-sized neighborhoods \cite{manderick1989fine}. Thereby, individuals can only interact with their neighbors, which enhances the exploration of the search space through the induced slow diffusion of solutions across the population. On the other hand, exploitation is carried out inside each neighborhood \cite{alba2004solving}. Therefore, while in classical GAs the population is structured in a unique panmictic group, in cGAs the whole population is arranged over a grid (typically two-dimensionals), on which the aforementioned neighborhood relation is defined. Two are the most frequently used neighborhood structures: i) NEWS, linear5, or Von Neumann, in which the neighborhood of each individual is composed by its North (N), East (E), West (W), and South (S) individuals; and ii) C9, or Moore, in which the neighborhood is given by NW, N, NE, W, E, SW, S and SE individuals. We recommend \cite{alba2009cellular} for additional information about cellular grid structures, and \cite{luna2018addressing,afshar2018novel,nebro2009mocell} for an excerpt of different theoretical works and applications of this particular kind of evolutionary algorithms.

As pointed, in cGAs each individual can only interact with its assigned neighbors. Thus, the genetic crossover operates inside the neighborhoods, modifying each individual with one of its neighbors. Furthermore, newly generated individuals are not introduced in the population. On the contrary, they replace the current individual upon the fulfillment of a given criterion (for example, an improvement in the fitness function). Additionally, two cGA types can be distinguished depending on the update policy of the population: synchronous cGA and asynchronous cGA. On the one hand, synchronous cGAs are characterized by implementing all the replacements in parallel. On the other hand, in asynchronous cGAs individuals are sequentially updated, thus overriding any need for auxiliary populations and adapting faster to the newly generated genetic material. These are the main reasons why we have chosen this second scheme for our research work.

\section{Multifactorial Cellular Genetic Algorithm}\label{sec:MFCGA}

As we have been identified previously, the four pillars on which the operation of the MFEA is based are unified representation, assortative mating, selective evaluation, and scalar fitness based selection. These concepts have been embraced and reformulated in this work to yield the workflow of the proposed MFCGA shown in Algorithm \ref{alg:MFCGA}, which are inspired by both MFEA and cGAs.
\begin{algorithm}[hb]
	 \SetAlgoLined
	 \DontPrintSemicolon
		Randomly generate a population of $P$ individuals\;
		Evaluate each generated individual for the $K$ problems\;
		Calculate the skill factor ($\tau^p$) of each individual $\mathbf{x}_p^\prime$\;
		Let $\mathbf{X}_p^{\circledast}$ denote the set of neighbors of $\mathbf{x}_p^\prime$\;
		\While{termination criterion not reached}{
		    \For{$p=1,\ldots,P$}{
        		Randomly choose a neighbor $\mathbf{x}_j$ from $\mathbf{X}_p^{\circledast}$\; 
        		$\mathbf{x}_p^{crossover}\gets\texttt{crossover}(\mathbf{x}_p^\prime,\mathbf{x}_j)$\;
        		$\mathbf{x}_p^{mutation}\gets\texttt{mutation}(\mathbf{x}_p^\prime)$\;
    		    Evaluate $\mathbf{x}_p^{crossover}$ and $\mathbf{x}_p^{mutation}$ for $\tau^p$\;
    		    $\mathbf{x}_p^\prime \gets \texttt{best}(\mathbf{x}_p^\prime,\mathbf{x}_p^{crossover},\mathbf{x}_p^{mutation})$\;
    		    Update $\varphi^p$ and $\tau^p$ of the evolved $\mathbf{x}_p^\prime$\;
    		}
		}
		Return the best individual for each task $T_k$\;
   \caption{Pseudocode of the proposed MFCGA}
	 \label{alg:MFCGA}
\end{algorithm}

First, as unified representation, the same philosophy and encoding as in the case of the MFEA has been used. Regarding the genetic operators, they are based on the classical evolutionary crossover and mutation procedures: at every generation, each individual $\mathbf{x}_p^\prime$ goes through these two phases (without using any crossover or mutation probabilities), producing two new individuals: $\mathbf{x}_p^{crossover}$ and $\mathbf{x}_p^{mutation}$. The first of these newly created individuals is the result of mating $\mathbf{x}_p^\prime$ with a randomly chosen neighbor $\mathbf{x}_j$ from the cellular neighborhood $\mathbf{X}_p^{\circledast}$ of $\mathbf{x}_p^\prime$. Correspondingly, the mutation operator applied to $\mathbf{x}_p^\prime$ gives rise to $\mathbf{x}_p^{mutation}$.

Once $\mathbf{x}_p^{crossover}$ and $\mathbf{x}_p^{mutation}$ have been generated, their quality is evaluated by using the same selective evaluation described in Subsection \ref{sec:MFEA}. In this way, we ensure that MFGCA is as computationally efficient as MFEA. It should be mentioned here that both $\mathbf{x}_p^{crossover}$ and $\mathbf{x}_p^{mutation}$ are evaluated for task $T_{\tau^p}$, where $\tau^p$ is the skill factor of $\mathbf{x}_p^\prime$. This implies a significant difference with respect to MFEA, since in MFGCA individuals are devoted to the optimization of the same single task along the whole execution, not changing at all from one task to another. Moreover, the first complete evaluation and sorting of the population, based on the factorial rank and scalar factor, ensures the equilibrium between the population, allocating a similar number of individuals to each of the tasks.

A final aspect of the proposed MFCGA is the local improvement selection mechanism, by which the newly generated $\mathbf{x}_p^{crossover}$ or $\mathbf{x}_p^{mutation}$ can only substitute their parent $\mathbf{x}_p^\prime$. In fact, the individual that survives to the next generation is the best one among $\mathbf{x}_p^\prime$, $\mathbf{x}_p^{crossover}$ and $\mathbf{x}_p^{mutation}$. The other two produced individuals are automatically discarded.

\section{Experimental Setup and Results}\label{sec:exp}

We proceed by describing an experimentation conducted for comparing both MFEA and MFCGA solvers, properly analyzing the genetic transfer within MFCGA and examining the synergies between the chosen tasks. As mentioned previously, the experimentation has been done over the well-known TSP \cite{bellmore1968traveling}. Since its inception, the TSP has become one of the most popular benchmark problems for the performance assessment of discrete optimization algorithms, from traditional meta-heuristics such as GAs \cite{grefenstette1985genetic} or Ant Colony Optimization \cite{dorigo1997ant}, to more recently introduced bio-inspired solvers, such as the Firefly Algorithm \cite{kumbharana2013solving}, Bat Algorithm \cite{osaba2016improved}, or the Water Cycle Algorithm \cite{osaba2018discrete}, among others. In the context of the present study, our main objective is not to find the optimal solution to the TSP problems under consideration. Instead, we aim to statistically compare the performance of both MFEA and MFCGA using same problem instances and conditions.

Specifically, The performance of the developed MFEA and MFCGA has been gauged over 15 different combinations (\emph{test cases}) of the Krolak/Felts/Nelson set of TSP instances contained in the renowned TSPLIB repository \cite{TSPLib}. It is important to highlight that these cases have been selected not only because of their wide acceptance by the community, but also since the different levels of genetic complementarities in their structure. This complementarity is measured using the percentage of nodes that instances share between them. Thus, our intention is to explore the impact of this complementarities in the genetic exchange inherent to EM schemes. In Table \ref{tab:similarity}, a summary of genetic complementarities is shown for all the datasets considered in the experimentation.
\begin{table}[ht!]
	\centering
	\caption{Summary of genetic complementarities for all the datasets employed in the experimentation}	
	\resizebox{\columnwidth}{!}{
		\begin{tabular}{ccccccccc}
			\toprule
			Instance & kroA100 & kroB100 & kroC100 & kroD100 & kroE100 & kroA150 & kroA200 &  kroB150\\
			\midrule
			kroA100 & \cellcolor{gray!25} & 1\% &  2\% & 1\% & 2\% & \textbf{80\%} & \textbf{66\%} & 1\%\\
			kroB100 & \cellcolor{gray!25} & \cellcolor{gray!25} & 1\% & 2\% & 1\% & 1\% & 0\% & 0\%\\
			kroC100 & \cellcolor{gray!25} & \cellcolor{gray!25} & \cellcolor{gray!25} & 1\% & 1\% & 1\% & \textbf{66\%} & \textbf{80\%}\\ 
			kroD100 & \cellcolor{gray!25} & \cellcolor{gray!25} & \cellcolor{gray!25} & \cellcolor{gray!25} & 1\% & 1\% & 2\% & 1\%\\
			kroE100 & \cellcolor{gray!25} & \cellcolor{gray!25} & \cellcolor{gray!25} & \cellcolor{gray!25} &\cellcolor{gray!25}  & \textbf{40\%} & 0\% & \textbf{40\%}\\ 
			kroA150 & \cellcolor{gray!25} & \cellcolor{gray!25} & \cellcolor{gray!25} & \cellcolor{gray!25} &\cellcolor{gray!25} & \cellcolor{gray!25} & \textbf{57\%} & 1\%\\
			kroA200 & \cellcolor{gray!25} & \cellcolor{gray!25} & \cellcolor{gray!25} & \cellcolor{gray!25} &\cellcolor{gray!25} & \cellcolor{gray!25} & \cellcolor{gray!25} & \textbf{57\%}\\ \bottomrule
		\end{tabular}
	}
	\label{tab:similarity}
\end{table}

Each of the 15 multitasking test cases implies that the modeled approaches should solve all the tasks assigned to that scenario. As shown in Table \ref{tab:testCases}, 10 of these test cases are comprised by four TSP instances, 4 are composed by 6 TSP instances, and the last one contemplates the resolution of all the 8 TSP problems under consideration. Two have been the main reasons of building these tests cases: i) to ensure the heterogeneity and variety of the configurations, meaning that each TSP instance is part of exactly the same number of test cases; and ii) to examine how the genetic synergies depicted in Table \ref{tab:similarity} are exploited during the search process by the proposed MFCGA approach.
\begin{table}[ht!]
    \centering
    \caption{Summary of the 15 test cases built for the experimentation}
    \resizebox{\columnwidth}{!}{
        \begin{tabular}{cL{7cm}}
        	\toprule
            Test Case & Tasks involved\\ \midrule
            \texttt{TC\_4\_1} & kroA100, kroA150, kroA200, kroC100\\ 
            \texttt{TC\_4\_2} & kroB100, kroB150, kroD100, kroE100\\ 
            \texttt{TC\_4\_3} & kroA100, kroA150, kroD100, kroE100\\ 
            \texttt{TC\_4\_4} & kroA200, kroC100, kroB100, kroB150\\ 
            \texttt{TC\_4\_5} & kroA100, kroA200, kroB100, kroD100\\ 
            \texttt{TC\_4\_6} & kroA150, kroC100, kroB150, kroE100\\ 
            \texttt{TC\_4\_7} & kroA100, kroA150, kroB100, kroB150\\ 
            \texttt{TC\_4\_8} & kroA200, kroC100, kroD100, kroE100\\ 
            \texttt{TC\_4\_9} & kroA100, kroC100, kroB100, kroD100\\ 
            \texttt{TC\_4\_10} & kroA150, kroA200, kroB150, kroE100\\ 
            \texttt{TC\_6\_1} & kroA100, kroA150, kroA200, kroB100, kroC100, kroB150\\ 
            \texttt{TC\_6\_2} & kroA200, kroB100, kroC100, kroB150, kroD100, kroE100\\ 
            \texttt{TC\_6\_3} & kroA100, kroA150, kroA200, kroB150, kroD100, kroE100\\ 
            \texttt{TC\_6\_4} & kroA100, kroA150, kroB100, kroC100, kroD100, kroE100\\ 
            \texttt{TC\_8} & kroA100, kroA150, kroA200, kroB100, kroC100, kroB150, kroD100, kroE100\\ \bottomrule
        \end{tabular}
    }
    \label{tab:testCases}
\end{table}

Regarding the algorithmic setup, we have used similar parameters and the same operators for all the implemented algorithms to ensure a fair comparison. For the sake of reproducibility of the presented results, the parameterization used for both MFEA and MFCGA are listed in Table \ref{tab:Parametrization}. For this parameterization, studies focused on cGAs and MFEA have been used as inspiration \cite{alba2004solving,yuan2016evolutionary}, along with the methodological guidelines given in \cite{osaba2018good}. Accordingly, results are reported on the basis of $20$ independent runs for every test case to inspect the statistical significance of eventual performance gaps. In addition, both MFEA and MFCGA are stopped after $500\cdot 10^3$ objective function evaluations. All experiments have been executed on an Intel Xeon E5–2650 v3 2.30 GHz processor with 32 GB RAM.
\begin{table}[h!]
	\centering
	\caption{Parametrization of MFEA and MFCGA}
	\resizebox{0.7\columnwidth}{!}{
		\begin{tabular}{lll}
			\toprule Parameter & MFEA & MFCGA\\
			\midrule
			Population size & \multicolumn{2}{c}{200} \\ 
			$\texttt{crossover}(\cdot)$ & \multicolumn{2}{c}{Order crossover \cite{davis1985job}} \\ 
			$\texttt{mutation}(\cdot)$ & \multicolumn{2}{c}{2-opt}\\ 
			Crossover probability & 0.9 & 1.0 \\ 
			Mutation probability & 0.1 & 1.0 \\ 
			Type of grid & \cellcolor{gray!25} & Moore \\ 
			\bottomrule
		\end{tabular}
	}
	\label{tab:Parametrization}
\end{table}

A Java implementation of the MFCGA has been made publicly available in \textit{https://git.code.tecnalia.com/aritz.martinez/mfcga}, together with the scripts that generate the results next discussed.

\subsection{Results and Discussion}\label{sec:exp_res}

We begin our discussion by analyzing Table \ref{tab:testCasesSummary}, which summarizes graphically the comparisons between the outcomes obtained by both MFCGA and MFEA in all the 15 test cases described above. Specifically, an orange circle $\tikzcircle[fill=orange]{3pt}$ indicates that MFCGA outperforms MFEA in terms of fitness average for a given TSP problem instance. On the other hand, the gray circle $\tikzcircle[fill=gray]{3pt}$ denotes that MFEA has reached better average outcomes. Let us take $\texttt{TC\_4\_2}$ as an example: in this case, and considering the order of instances within the test case provided in Table \ref{tab:testCases}, we can observe that MFCGA performs better in kroB100, kroB150, and kroE100, while MFEA dominates only in the kroE100 instance. By extrapolating this analysis to the remaining content of the table, we conclude that MFCGA elicits a better performance for tackling these test cases, outperforming MFEA in all but six problem instances. It is also important to underscore that for $\texttt{TC\_8}$, MFCGA attains better performance scores in all its compounding 8 TSP instances.
\begin{table}[ht!]
    \centering
    \caption{Comparison of the results for the 15 test cases ($\tikzcircle[fill=orange]{3pt}$: MFCGA outperforms MFEA; $\tikzcircle[fill=gray]{3pt}$: MFEA outperforms MFCGA).}
    \resizebox{0.65\columnwidth}{!}{
        \begin{tabular}{cL{3.5cm}}
            \toprule Test Case & MFGCA versus MFEA\\ \midrule
            \texttt{TC\_4\_1} & \tikzcircle[fill=orange]{4pt}-\tikzcircle[fill=orange]{4pt}-\tikzcircle[fill=orange]{4pt}-\tikzcircle[fill=orange]{4pt} \\ 
            \texttt{TC\_4\_2} & \tikzcircle[fill=orange]{4pt}-\tikzcircle[fill=orange]{4pt}-\tikzcircle[fill=gray]{4pt}-\tikzcircle[fill=orange]{4pt} \\ 
            \texttt{TC\_4\_3} & \tikzcircle[fill=orange]{4pt}-\tikzcircle[fill=orange]{4pt}-\tikzcircle[fill=orange]{4pt}-\tikzcircle[fill=orange]{4pt} \\ 
            \texttt{TC\_4\_4} & \tikzcircle[fill=orange]{4pt}-\tikzcircle[fill=gray]{4pt}-\tikzcircle[fill=orange]{4pt}-\tikzcircle[fill=orange]{4pt}\\ 
            \texttt{TC\_4\_5} & \tikzcircle[fill=orange]{4pt}-\tikzcircle[fill=orange]{4pt}-\tikzcircle[fill=orange]{4pt}-\tikzcircle[fill=orange]{4pt}\\ 
            \texttt{TC\_4\_6} & \tikzcircle[fill=orange]{4pt}-\tikzcircle[fill=orange]{4pt}-\tikzcircle[fill=orange]{4pt}-\tikzcircle[fill=orange]{4pt}\\ 
            \texttt{TC\_4\_7} & \tikzcircle[fill=orange]{4pt}-\tikzcircle[fill=gray]{4pt}-\tikzcircle[fill=orange]{4pt}-\tikzcircle[fill=orange]{4pt}\\ 
            \texttt{TC\_4\_8} & \tikzcircle[fill=orange]{4pt}-\tikzcircle[fill=orange]{4pt}-\tikzcircle[fill=orange]{4pt}-\tikzcircle[fill=gray]{4pt}\\ 
            \texttt{TC\_4\_9} & \tikzcircle[fill=orange]{4pt}-\tikzcircle[fill=orange]{4pt}-\tikzcircle[fill=orange]{4pt}-\tikzcircle[fill=orange]{4pt}\\ 
            \texttt{TC\_4\_10} & \tikzcircle[fill=orange]{4pt}-\tikzcircle[fill=orange]{4pt}-\tikzcircle[fill=orange]{4pt}-\tikzcircle[fill=orange]{4pt}\\ 
            \texttt{TC\_6\_1} & \tikzcircle[fill=orange]{4pt}-\tikzcircle[fill=orange]{4pt}-\tikzcircle[fill=orange]{4pt}-\tikzcircle[fill=orange]{4pt}-\tikzcircle[fill=gray]{4pt}-\tikzcircle[fill=orange]{4pt}\\ 
            \texttt{TC\_6\_2} & \tikzcircle[fill=orange]{4pt}-\tikzcircle[fill=orange]{4pt}-\tikzcircle[fill=orange]{4pt}-\tikzcircle[fill=orange]{4pt}-\tikzcircle[fill=orange]{4pt}-\tikzcircle[fill=gray]{4pt}\\ 
            \texttt{TC\_6\_3} & \tikzcircle[fill=orange]{4pt}-\tikzcircle[fill=orange]{4pt}-\tikzcircle[fill=orange]{4pt}-\tikzcircle[fill=orange]{4pt}-\tikzcircle[fill=orange]{4pt}-\tikzcircle[fill=orange]{4pt}\\ 
            \texttt{TC\_6\_4} & \tikzcircle[fill=orange]{4pt}-\tikzcircle[fill=orange]{4pt}-\tikzcircle[fill=orange]{4pt}-\tikzcircle[fill=orange]{4pt}-\tikzcircle[fill=orange]{4pt}-\tikzcircle[fill=orange]{4pt}\\ 
            \texttt{TC\_8} & \tikzcircle[fill=orange]{4pt}-\tikzcircle[fill=orange]{4pt}-\tikzcircle[fill=orange]{4pt}-\tikzcircle[fill=orange]{4pt}-\tikzcircle[fill=orange]{4pt}-\tikzcircle[fill=orange]{4pt}-\tikzcircle[fill=orange]{4pt}-\tikzcircle[fill=orange]{4pt}\\ \bottomrule
        \end{tabular}
    }
    \label{tab:testCasesSummary}
\end{table}

Table \ref{tab:bestSolutions} exemplifies the process by which the above results have been produced for the \texttt{TC\_8} test case. In this table we depict, for each TSP instance in the test case, the average, best and standard deviation of the fitness value achieved by MFEA and MFCGA computed over 20 independent runs. Additionally, we also represent the known optima for each instance. It is straightforward to note that MFCGA clearly outperforms MFEA in terms of average results. Furthermore, regarding the best solution found over the 20 independent runs, MFCGA also dominates the benchmark, obtaining a better performance in 5 out of the 8 cases. Finally, it is also interesting to notice that the difference between the known optima and the average outcomes obtained by MFCGA ranges between $3.8$\% and $4.7$\% in problem instances with 100 nodes, and between $4.7$\% and $9.3$\% for problems of larger size.

In order to verify the statistical significance between the results returned by MFCGA and MFEA, the Wilcoxon Rank-Sum test has been applied to their fitness outcomes. The confidence interval has been set at $95$\%. For properly building this statistical test, we have compared the results reached in all the 8 datasets separately, depicting graphically the outcomes of these Wilcoxon Rank-Sum tests in the last row of Table \ref{tab:bestSolutions}. In this row, an orange circle $\tikzcircle[fill=orange]{3pt}$ means that MFCGA outperforms MFEA with statistical significance. On the other hand, the gray circle $\tikzcircle[fill=gray]{3pt}$ denotes that there is no enough evidence to claim that the improvement is statistically relevant. As a summary of all these tests, the obtained average $z$-value is $-1.96$, with an average $p$-value equal to $0.04888$. Taking into account the critical $z_c$ value is equal to $-1.64$, and since $-1.96<-1.64$ and $0.04888<0.05$, these results support the significance of the difference at $95$\% confidence level. Therefore, the difference is significant at this confidence level, thereby concluding that MFCGA is statistically better than MFEA for this test case.
\begin{table}[ht!]
	\centering
	\caption{Results obtained by MFCGA and MFEA for the 8 instances in \texttt{TC\_8}, and graphical results of the Wilcoxon Rank-Sum test.}
	\resizebox{\columnwidth}{!}{
		\begin{tabular}{C{1.3cm}cccccccc}
			\toprule
			Method & kroA100 & kroA150 & kroA200 & kroB100 & kroC100 & kroB150 & kroD100 &  kroE100\\ 
			\midrule
			\multirow{3}{*}{MFCGA} & \textbf{22099.1} & \textbf{28588.1} & \textbf{32109.0} & \textbf{23168.9} & \textbf{21494.7} & \textbf{27780.5} & \textbf{22257.7} & \textbf{23069.4}\\ 
			& 21746.0 & \textbf{27893.0} & \textbf{31162.0} & 22815.0 & \textbf{20852.0} & \textbf{27307.0} & 21648.0 & \textbf{22587.0}\\ 
			& 203.89 & 394.96 & 547.66 & 249.66 & 337.44 & 207.34 & 398.46 & 211.52\\ \midrule
			\multirow{3}{*}{MFEA} & 22404.6 & 28817.0 & 32769.8 & 23790.0 & 21956.3 & 28512.0 & 22713.3 & 23239.3 \\
			& \textbf{21460.0} & 28385.0 & 31856.0 & \textbf{22330.0} & 21157.0 & 27394.4 & \textbf{21539.0} & 22607.0 \\
			& 703.53 & 299.44 & 596.42 & 609.08 & 739.15 & 788.14 & 548.83 & 533.68 \\ \midrule
			Optima & 21282.0 &  26524.0 & 29368.0 & 22141.0 & 20749.0 & 26524.0 & 21294.0 & 22068.0 \\ 
			\midrule
			Wilcoxon rank-sum test & \tikzcircle[fill=gray]{5pt} & \tikzcircle[fill=orange]{5pt} & \tikzcircle[fill=orange]{5pt} & \tikzcircle[fill=orange]{5pt} & \tikzcircle[fill=orange]{5pt} & \tikzcircle[fill=orange]{5pt} & \tikzcircle[fill=orange]{5pt} & \tikzcircle[fill=gray]{5pt} \\ \bottomrule
		\end{tabular}
	}
	\label{tab:bestSolutions}
\end{table}

\subsection{Analysis of the Genetic Transfer between Tasks}\label{sec:exp_gen}

We now analyze the genetic transfer across the 8 TSP tasks considered in the complete experimentation, focusing on our proposed MFCGA. The main objective with this study is i) to get a glimpse of the positive knowledge transfer among problem instances; ii) to discover synergies between them; and iii) to empirically gauge inter-task interactions occurred along the 20 repetitions of \texttt{TC\_8}. We have chosen this test case since it is the one in which the 8 TSP tasks are optimized jointly.

It should be pointed here that the novel MFCGA presented in this paper is especially interesting for analyzing the genetic transfer held through the algorithm execution. This is so due to the replacement strategy employed in MFCGA. In our method, an individual $\mathbf{x}_p^\prime$ of the population is replaced if and only if any of the individuals generated through the \texttt{crossover} ($\mathbf{x}_p^{crossover}$) and \texttt{mutation} ($\mathbf{x}_p^{mutation}$) operators outperform $\mathbf{x}_p^\prime$ in terms of its best performing task (i.e. its skill factor). Thus, if $\mathbf{x}_p^{crossover}$ replaces $\mathbf{x}_p^\prime$, a positive transfer of genetic material has occurred from $\mathbf{x}_j$ to $\mathbf{x}_p^\prime$ (we refer to Algorithm \ref{alg:MFCGA} and Section \ref{sec:MFCGA} for notation details). In the context of the TSP, this transfer is realized through the direct insertion of part of the neighboring solution $\mathbf{x}_j$ into $\mathbf{x}_p^\prime$, which can be conceived as a positive contribution of task $\tau^j$ to task $\tau^p$.

\begin{figure}[h!]
	\centering
	\includegraphics[width=0.85\hsize]{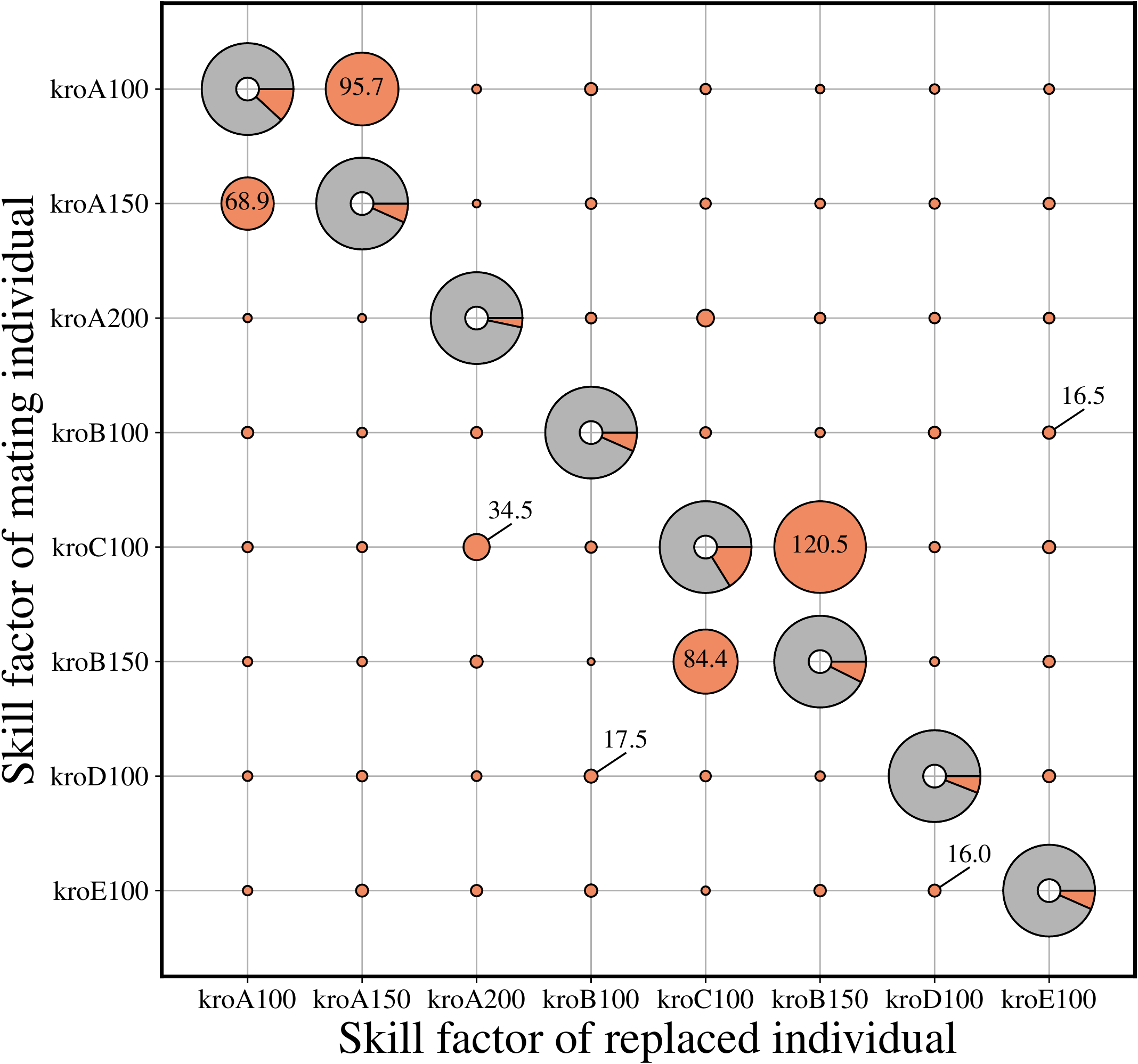}
	\caption{Average intensities of genetic transfer between TSP instances.}
	\label{fig:matrix_influence}
\end{figure}

Bearing the above explanation in mind, Figure \ref{fig:matrix_influence} represents the number of positive genetic transfer episodes (through the \texttt{crossover} operator) between every pair of TSP tasks. The radius of every orange circle in this plot is proportional to the average number of times per execution in which an individual having the skill factor indicated in the column label has exchanged some of its genetic material with an individual whose skill factor is given in the row label. Moreover, circles located in the diagonal represent the sum of all the inter-task (orange portion) and intra-task exchanges (gray portion), the latter quantifying the genetic transfer between individuals featuring the same skill factor.

Two main conclusions can be drawn after analyzing this figure. The first one is the confirmation of synergy in three different pairs of TSP instances, namely, \{kroA100, kroA150\}, \{kroC100, kroA200\} and \{kroC100, kroB150\}. Thus, the genetic transfer between these tasks positively contributes to the multi-task search process. The second conclusion is that for the rest of task pairs, the intensity of genetic material exchange is almost nonexistent. This fact unveils that transfers between these tasks can be considered as negative \cite{bonilla2008multi}, and that they do not contribute for the search process. These findings a priori contradict the information depicted in Table \ref{tab:similarity}, in which we summarized the genetic complementarity of the 8 tasks. For instance, we saw that tasks such as kroA100 and kroA200 have a high level of complementarity among them (66\% as per the table). However, the inter-task interaction for this pair depicted in Figure \ref{fig:matrix_influence} is practically nonexistent. Similar conclusions hold for other task pairs, such as \{kroA150, kroA200\} or \{kroA200, kroB150\}. This contradiction collides with some previously published studies \cite{gupta2015multifactorial}. In fact, by analyzing the correlation in the landscapes of the aforementioned task pairs, we can confirm that the so-called partial domain overlap exists \cite{gupta2017insights}. This statement is confirmed since the domains of task pairs \{kroA100, kroA150\}, \{kroA150, kroA200\} and \{kroA200, kroB150\} partially overlap, existing a subset of features that are common to both tasks of every pair. 

In order to shed light on this unexpected mismatch, a deeper analysis of the considered 8 TSP instances has been made. However, in this case we focus our attention on the correlation among the best known solutions of such instances. Measuring the distance between best solutions has been previously proposed in recent works on continuous problems \cite{da2017evolutionary,zhou2018study}. We summarize in Table \ref{tab:bestSolutionsSim} the genetic complementarities in the optimal solutions of the 8 TSP tasks in use. Cells corresponding to task pairs that have shown a higher inter-task genetic transfer in Figure \ref{fig:matrix_influence} have been highlighted in orange. As shown in this table, the best known solutions of \{kroA100, kroA150\}, \{kroC100, kroA200\} and \{kroC100, kroB150\} present a partial degree of intersection, which means that \textit{the global optima of the two tasks are identical in the unified search space with respect to a subset of variables only, and different with respect to the remaining variables} \cite{da2017evolutionary}. At the same time, these three pairs are the ones that evince a higher intensity of interaction in the conducted experiments.
\begin{table}[ht!]
    \centering
    \caption{Genetic complementarities among the best known solutions of the TSP instances utilized in the experimentation}
    \resizebox{\columnwidth}{!}{
        \begin{tabular}{ccccccccc}
        	\toprule
            Instance & kroA100 & kroB100 & kroC100 & kroD100 & kroE100 & kroA150 & kroA200 &  kroB150\\ 
            \midrule
            kroA100 & \cellcolor{gray!25} & 0\% & 0\% & 0\% & 0\% & \cellcolor{orange!25}\textbf{32\%} & \textbf{5\%} & 0\%\\ 
            kroB100 & \cellcolor{gray!25} & \cellcolor{gray!25} & 0\% & 0\% & 0\% & 0\% & 0\% & 0\%\\ 
            kroC100 & \cellcolor{gray!25} & \cellcolor{gray!25} & \cellcolor{gray!25} & 0\% & 0\% & 0\% &  \cellcolor{orange!25}\textbf{21\%} &  \cellcolor{orange!25}\textbf{10\%}\\ 
            kroD100 & \cellcolor{gray!25} & \cellcolor{gray!25} & \cellcolor{gray!25} & \cellcolor{gray!25} & 0\% & 0\% & 0\% & 0\%\\ 
            kroE100 & \cellcolor{gray!25} & \cellcolor{gray!25} & \cellcolor{gray!25} & \cellcolor{gray!25} &\cellcolor{gray!25}  & \textbf{3\%} & 0\% & \textbf{2\%}\\ 
            kroA150 & \cellcolor{gray!25} & \cellcolor{gray!25} & \cellcolor{gray!25} & \cellcolor{gray!25} &\cellcolor{gray!25} & \cellcolor{gray!25} & \textbf{3\%} & 0\%\\ 
            kroA200 & \cellcolor{gray!25} & \cellcolor{gray!25} & \cellcolor{gray!25} & \cellcolor{gray!25} &\cellcolor{gray!25} & \cellcolor{gray!25} & \cellcolor{gray!25} & \textbf{8\%}\\ \bottomrule
        \end{tabular}
    }
    \label{tab:bestSolutionsSim}
\end{table}

This last analysis leads to the ultimate finding of our experimentation: the confirmation that for the TSP, positive inter-task genetic transfer is likely to happen among pairs of optimization tasks in which the degree of intersection in their best solution is, at least, partial. Specifically, our experiments elucidate that the degree of intersection should be above 10\% for the transfer between tasks to be beneficial. In other words, the raw complementarity in the structure of the TSP scenario is irrelevant for the genetic transfer. For this reason, we conclude that TSP instances which do not partially share a fraction of their best solutions are prone to negative inter-task interactions in EM environments.

\section{Conclusions and Future Work}\label{sec:conc}

This work has elaborated on the design, implementation and performance assessment of a novel Multifactorial Cellular Genetic Algorithm (MFCGA) for Evolutionary Multitasking. Our proposed meta-heuristic approach is inspired by the well-known MFEA, and the influential cellular Genetic Algorithm (cGA). Specifically, the meta-heuristic search strategy relies on a neighborhood relationship induced on a grid arrangement of the individuals of the population, which restricts the coverage of the evolutionary crossover operator. For assessing the quality of our method, we have compared the performance of the MFCGA to that of MFEA along 15 tests cases comprising 8 different TSP instances. The obtained experimental outcomes support the preliminary conclusion that MFCGA is a promising method for solving EM environments. An equally important contribution of this work is the inter-task genetic transfer analysis conducted over the MFCGA, aimed at uncovering synergistic relationships among TSP instances that are exploited over the search process. Our main conclusion on this regard is that the genetic exchange can be positive whenever tasks present a minimum degree of intersection in the structure of their best solutions.

We plan to devote further efforts in a manifold of interesting research paths rooted on this initial study. In the short term, we will continue using the TSP as benchmarking problem using larger instances and test cases, targeting to assess the scalability of the developed MFCGA. Furthermore, additional search mechanisms for our method will be investigated and tested, such as heuristic local search methods or alternative survivor strategies. In the longer term, we will explore the application of the MFCGA to other fields \cite{precup2019nature} and additional discrete optimization problems, such as the vehicle routing \cite{caceres2015rich}, or community detection problems \cite{pizzuti2017evolutionary,osaba2019community}. In those cases, a similar analysis of the intra-task genetic transfer will be undertaken, possibly by resorting to other means for computing the similarity between solutions. Finally, a closer look will be taken at adaptive means to efficiently exploit synergies between solutions, by potentially optimizing the distribution of individuals over the grid according to such intra-task relationships. We will also try to adapt additional existing methods to this field, such as the firefly algorithm \cite{yang2010firefly}, bat algorithm \cite{yang2010new} or grey wolf optimizer \cite{mirjalili2014grey,precup2016grey}. 

\section*{Acknowledgments}

Eneko Osaba, Aritz D. Martinez, Jesus L. Lobo and Javier Del Ser would like to thank the Basque Government for its funding support through the EMAITEK and ELKARTEK programs. Javier Del Ser receives funding support from the Consolidated Research Group MATHMODE (IT1294-19) granted by the Department of Education of the Basque Government.

\bibliographystyle{./IEEEtran}
\bibliography{./IEEEexample}

\end{document}